\documentclass[10pt,twocolumn,letterpaper]{article}

\usepackage{iccv}
\usepackage{times}
\usepackage{epsfig}
\usepackage{graphicx}
\usepackage{amsmath}
\usepackage{amssymb}
\usepackage{multirow}
\usepackage{bbding}

\usepackage[pagebackref=true,breaklinks=true,letterpaper=true,colorlinks,bookmarks=false]{hyperref}

 \iccvfinalcopy 

\def\iccvPaperID{3019} 

\ificcvfinal\fi

\begin{document}


\title{Temporal Contrastive Graph Learning for Video Action Recognition and Retrieval}
\author{Yang Liu\\
Sun-Yat-Sen University\\
{\tt\small liuy856@mail.sysu.edu.cn}
\and
Keze Wang\\
DarkMatter AI\\
{\tt\small kezewang@gmail.com}
\and
Haoyuan Lan\\
Sun-Yat-Sen University\\
{\tt\small lanhy5@mail2.sysu.edu.cn}
\and
Liang Lin\\
Sun-Yat-Sen University\\
{\tt\small linliang@ieee.org}
}
\maketitle
\ificcvfinal\thispagestyle{empty}\fi
\begin{abstract}\vspace{-10pt}
Attempt to fully discover the temporal diversity and chronological characteristics for self-supervised video representation learning, this work takes advantage of the temporal dependencies within videos and further proposes a novel self-supervised method named Temporal Contrastive Graph Learning (TCGL). In contrast to the existing methods that ignore modeling elaborate temporal dependencies, our TCGL roots in a hybrid graph contrastive learning strategy to jointly regard the inter-snippet and intra-snippet temporal dependencies as self-supervision signals for temporal representation learning. To model multi-scale temporal dependencies, our TCGL integrates the prior knowledge about the frame and snippet orders into graph structures, i.e., the intra-/inter- snippet temporal contrastive graphs. By randomly removing edges and masking nodes of the intra-snippet graphs or inter-snippet graphs, our TCGL can generate different correlated graph views. Then, specific contrastive learning modules are designed to maximize the agreement  between nodes in different views. To adaptively learn the global context representation and recalibrate the channel-wise features, we introduce an adaptive video snippet order prediction module, which leverages the relational knowledge among video snippets to predict the actual snippet orders. Experimental results demonstrate the superiority of our TCGL over the state-of-the-art methods on large-scale action recognition and video retrieval benchmarks.
\end{abstract}
\vspace{-20pt}

\begin{figure}
\begin{center}
\includegraphics[scale=0.38]{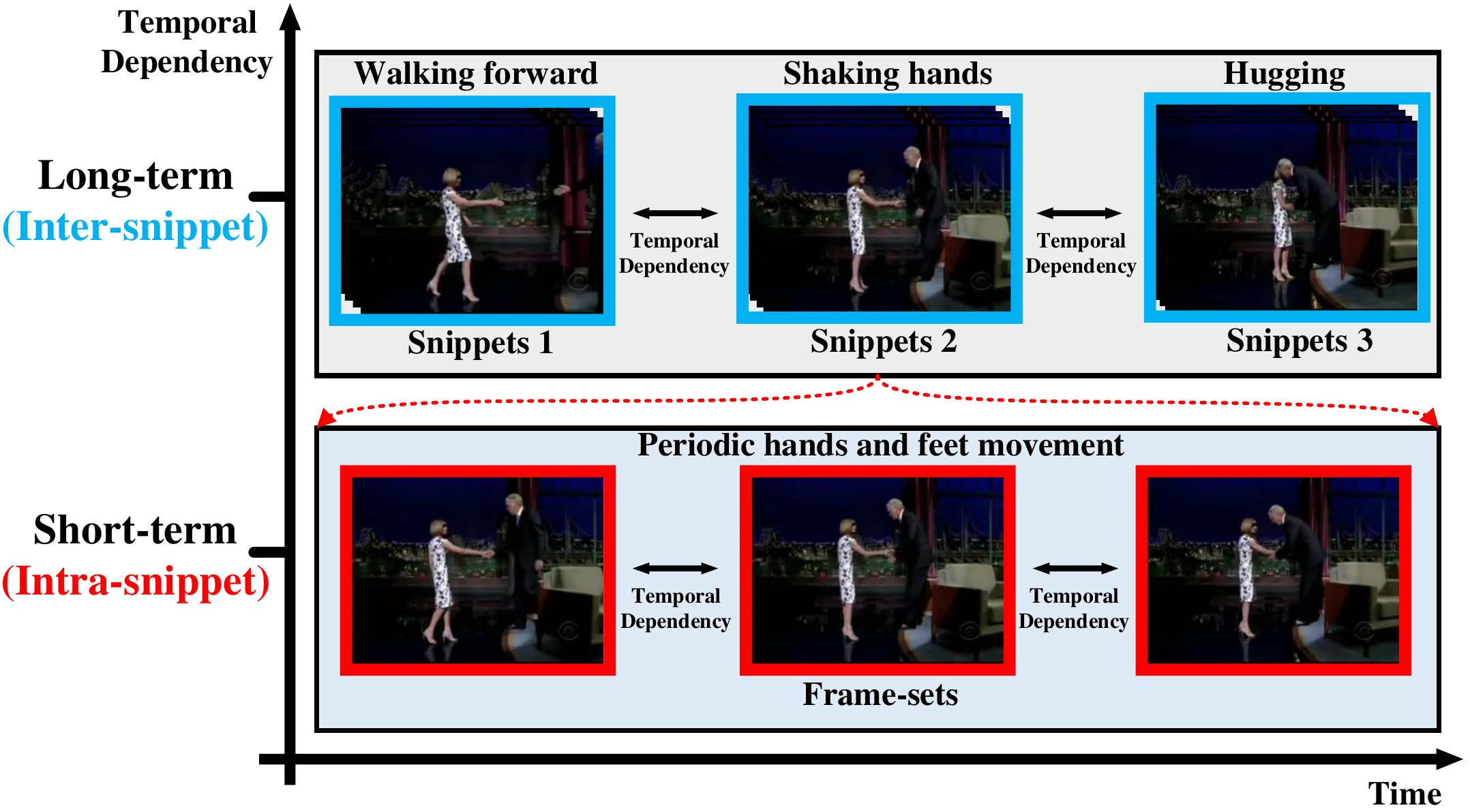}
\end{center}
  \vspace{-10pt}
   \caption{Illustration of the multi-scale temporal dependencies in videos. The action of handshaking contains the long-term  (inter-snippet) temporal dependencies of walking forward, shaking hands, and hugging, while it also includes the short-term (intra-snippet) temporal dependencies of periodic hands and feet movement. }
   \vspace{-15pt}
\label{Fig1}
\end{figure}

\section{Introduction}

Deep convolutional neural networks (CNNs) \cite{krizhevsky2012imagenet} have achieved state-of-the-art performance in many visual recognition tasks. This can be primarily attributed to the learned rich representation from well-trained networks using large-scale image/video datasets (\eg ImageNet \cite{deng2009imagenet}, Kinetics \cite{kay2017kinetics}, SomethingSomething \cite{goyal2017something}) with strong supervision information \cite{kim2019self}. However, annotating such large-scale data is laborious, expensive, and impractical, especially for complex data-based high-level tasks, such as video action understanding and video retrieval. To fully leverage the existing large amount of unlabeled data, self-supervised learning gives a reasonable way to utilize the intrinsic characteristics of unlabeled data to obtain supervisory signals, which has attracted increasing attention.

Different from image data that can be handled by defining proxy tasks (e.g., predicting relative positions of image patches \cite{doersch2015unsupervised}, solving  jigsaw puzzles \cite{noroozi2016unsupervised}, inpainting images \cite{pathak2016context}, and predicting the image color channel \cite{larsson2017colorization}) for self-supervised learning, video data additionally contains temporal information that can be leveraged to learn the supervisory signals. Recently, a variety of approaches have been proposed such as order verification \cite{misra2016shuffle,fernando2017self}, order prediction \cite{lee2017unsupervised, xu2019self,wang2020self}, speediness prediction \cite{benaim2020speednet,yao2020video}. However, all of these methods consider the temporal dependency only from a single scale (i.e., short-term or long-term) and ignore the multi-scale temporal dependencies, i.e., they extract either snippet-level or frame-level features via 2D/3D CNNs and neglect to integrate these features to model temporal dependencies at multiple time scales.

In this work, we argue that modeling multi-scale temporal dependencies is beneficial for various video classification tasks. Firstly, the recent neuroscience studies \cite{livingstone1988segregation,van1994neural, hans2020visual,cornsweet2012visual, milner2006visual} prove that the human visual system can perceive detailed motion information by capturing both long-term and short-term temporal dependencies. This has been inspired by several famous supervised learning methods (e.g., Nonlocal \cite{wang2018non}, PSANet \cite{zhao2018psanet}, GloRe \cite{chen2019graph}, and ACNet \cite{wang2019adaptively}). Secondly, an action usually consists of several temporal dependencies at both short-term and long-term timescales. As shown in Figure 1, the action of handshaking contains the long-term temporal dependencies (video snippets) of walking forward, shaking hands, and hugging, while it also includes the short-term temporal dependencies (frame-sets within a snippet) of periodic hands and feet movement. Randomly shuffling the frames or snippets cannot preserve the semantic content of the video. Actually, the short-term temporal dependencies within a video snippet is important especially for videos that contain strict temporal coherence, such as videos in SomethingSomething datasets \cite{goyal2017something}. Therefore, both short-term (e.g., intra-snippet) and long-term (e.g., inter-snippet) temporal dependencies are essential and should be jointly modeled to learn discriminative temporal representations for unlabeled videos.

Inspired by the convincing performance and high interpretability of graph convolutional networks (GCN) \cite{kipf2016semi, velivckovic2018graph, zhang2020deep}, several works \cite{wang2018non,wang2018videos,zhuo2019explainable,ji2020action,zhang2020temporal} were proposed to increase the temporal diversity of videos by using GCN in such a supervised learning fashion with large amounts of labeled data. Unfortunately, due to the lack of principles or guidelines to explore the intrinsic temporal knowledge of unlabeled video data, it is quite challenging to utilize GCN for self-supervised video representation learning.

Attempt to address this issue by regarding the jointly modeling the inter-snippet and intra-snippet temporal dependencies as guidelines, this work presents a novel self-supervised approach, named Temporal Contrastive Graph Learning (TCGL), targeting at learning the multi-scale temporal dependency knowledge within videos by guiding the video snippet order prediction in an adaptive manner. Specifically, a given video is sampled into several fixed-length snippets and then randomly shuffled. For each snippet, all the frames from this snippet are sampled into several fixed-length frame-sets. We utilize 3D convolutional neural networks (CNNs) as the backbone network to extract features for these snippets and frame-sets. To preserve both inter-snippet and intra-snippet temporal dependencies within videos, we propose graph neural network (GNN) structures with prior knowledge about the snippet orders and frame-set orders. The video snippets of a video and their chronological characteristics are used to construct the inter-snippet temporal graph. Similarly, the frame-sets within a video snippet and their chronological characteristics are leveraged to construct the intra-snippet temporal graph. Furthermore, we randomly remove edges and mask node features of the intra-snippet graphs or inter-snippet graphs to generate different correlated graph views. Then, specific contrastive learning modules are designed to enhance its discriminative capability for temporal representation learning. To learn the global context representation and recalibrate the channel-wise features adaptively, we propose an adaptive video snippet order prediction module, which leverages relational knowledge among video snippets to predict the actual snippet orders. The main contributions of the paper can be summarized as follows:
\begin{itemize}\setlength{\itemsep}{0pt }\setlength{\parsep}{0pt}\setlength{\parskip}{0pt}\vspace{-5pt}
\item Integrated with intra-snippet and inter-snippet temporal dependencies, we propose intra-snippet and inter-snippet temporal contrastive graphs to increase the temporal diversity among video frames and snippets in a graph contrastive self-supervised learning manner.
\item To learn the global context representation and recalibrate the channel-wise features adaptively for each video snippet, we propose an adaptive video snippet order prediction module, which employs the relational knowledge among video snippets to predict orders.
\item Extensive experiments on three networks and two downstream tasks show that the proposed method achieves state-of-the-art performance and demonstrate the great potential of the learned video representations.
\end{itemize}\vspace{-5pt}

The rest of the paper is organized as follows. We first review related works in Section 2, then the details of the proposed method are explained in Section 3. In Section 4, the implementation and results of the experiments are provided and analyzed. Finally, we conclude our works in Section 5.

\section{Related Work}
In this section, we will introduce the recent works on supervised and self-supervised video representation learning.
\subsection{Supervised Video Representation Learning}
\begin{figure*}
\begin{center}
\includegraphics[scale=0.37]{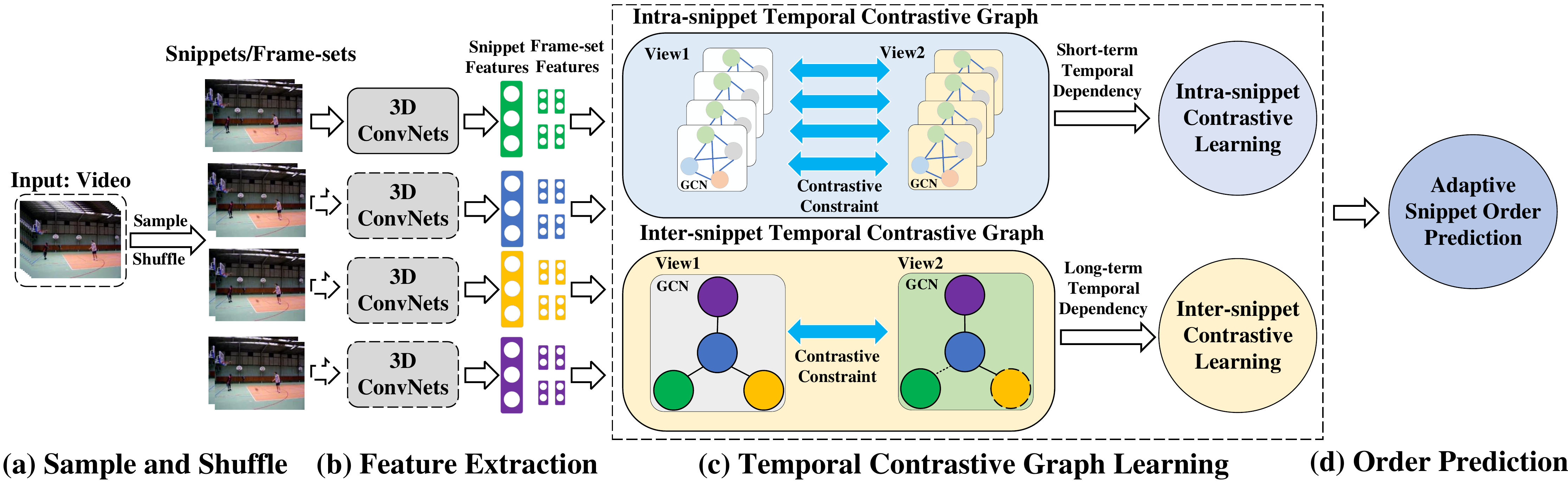}
\end{center}
  \vspace{-10pt}
   \caption{Overview of the Temporal Contrastive Graph Learning (TCGL) framework. (a) Sample and Shuffle: sample non-overlapping snippets for each video and randomly shuffle their orders. For each snippet, all the frames from this snippet are sampled into several fixed-length frame-sets. (b) Feature Extraction: use the 3D CNNs to extract the features for snippets and frame-sets. (c) Temporal Contrastive Graph Learning: intra-snippet and inter-snippet temporal contrastive graphs are constructed with the prior knowledge about the frame-set and snippet orders, see Figure \ref{Fig3} for more details. (d) Order Prediction: the learned snippet features from the temporal contrastive graph are adaptively forwarded through an adaptive snippet order prediction module to output the probability distribution over the possible orders. }
    \vspace{-10pt}
\label{Fig2}
\end{figure*}

For video representation learning, a large number of supervised learning methods have been received increasing attention. The methods include traditional methods \cite{laptev2005space,klaser2008spatio,wang2013dense,wang2013action,nguyen2014stap,peng2016bag,liu2018hierarchically} and deep learning methods \cite{simonyan2014two,tran2015learning,wang2015action,liu2018global,tran2018closer,wang2018temporal,zhou2018temporal,lin2019tsm,liu2019deep}. To model and discover temporal knowledge in videos, two-stream CNNs \cite{simonyan2014two} judged the video image and dense optical flow separately, then directly fused the class scores of these two networks to obtain the classification result. C3D \cite{tran2015learning} processed videos with a three-dimensional convolution kernel. Temporal Segment Networks (TSN) \cite{wang2018temporal} sampled each video into several segments to model the long-range temporal structure of videos. Temporal Relation Network (TRN) \cite{zhou2018temporal} introduced an interpretable network to learn and reason about temporal dependencies between video frames at multiple temporal scales. Temporal Shift Module (TSM) \cite{lin2019tsm} shifted part of the channels along the temporal dimension to facilitate information exchanged among neighboring frames. Although these supervised methods achieve promising performance in modeling temporal dependencies, they require large amounts of labeled videos for training an elaborate model, which is time-consuming and labor-intensive.

\subsection{Self-supervised Video Representation Learning}
Although there exists a large amount of videos, it may take a great effort to annotate  such massive data. Self-supervised learning generates various pretext tasks to leverage abundant unlabeled data. The learned model from pretext tasks can be directly applied to downstream tasks for feature extraction or fine-tuning. Specific contrastive learning methods have been proposed, such as the NCE \cite{gutmann2010noise}, MoCo \cite{he2020momentum}, BYOL \cite{grill2020bootstrap}, SimCLR \cite{DBLP:conf/icml/ChenK0H20}. To better model topologies, contrastive learning methods on graphs \cite{qiu2020gcc,zhu2020deep,hafidi2020graphcl} have also been attracted increasing attention.

For self-supervised video representation learning, how to effectively explore temporal information is important. Many existing works focus on the discovering of temporal information. Shuffle\&Learn \cite{misra2016shuffle} randomly shuffled video frames and trained a network to distinguish whether these video frames are in the right order or not. Odd-one-out Network \cite{fernando2017self} proposed to identify unrelated or odd video clips. Order prediction network (OPN) \cite{lee2017unsupervised} trained networks to predict the correct order of shuffled frames. VCOP \cite{xu2019self} used 3D convolutional networks to predict the orders of shuffled video clips. SpeedNet \cite{benaim2020speednet} designed a network to detect whether a video is playing at a normal rate or sped up rate. Video-pace \cite{wang2020self} utilized a network to identify the right paces of different video clips. In addition to focusing on the temporal dependency, Mas \cite{wang2019self} proposed a self-supervised learning method by regressing both motion and appearance statistics along spatial and temporal dimensions. ST-puzzle \cite{kim2019self} used space-time cubic puzzles to design pretext task. IIC \cite{tao2020self} introduced intra-negative samples by breaking temporal relations in video clips, and use these samples to build an inter-intra contrastive framework. Though the above works utilize temporal dependency or design specific pretext tasks for video self-supervised learning, the comprehensive temporal diversity and chronological characteristics are not fully explored. In our work, we build a novel inter-intra snippet graph structure to model multi-scale temporal dependencies, and produce self-supervision signals about video snippet orders contrastively.

\section{Temporal Contrastive Graph Learning}

In this section, we first give a brief overview of the proposed TCGL, shown in Figure \ref{Fig2}, which mainly consists of four stages. (1) Sample and shuffle, for each video, several snippets are uniformly sampled and shuffled. For each snippet,  all the frames from this snippet are sampled into several fixed-length frame-sets; (2) Feature extraction, 3D CNNs are utilized to extract features for these snippets and frame-sets, and all 3D CNNs share the same weights; (3) Temporal contrastive graph learning, we build two kinds of temporal contrastive graph structures (intra-snippet graph and inter-snippet graph) with the prior knowledge about the frame-set orders and snippet orders. To generate different correlated graph views for specific graphs, we randomly remove edges and mask node features of the intra-snippet graphs or inter-snippet graphs. Then, we design specific contrastive losses for both the intra-snippet and inter-snippet graphs to model multi-scale temporal dependencies. This can increase the temporal diversity of video representations; (4) Order prediction, the learned snippet features from the temporal contrastive graphs are adaptively forwarded through an adaptive snippet order prediction module to output the probability distribution over the possible orders.

\begin{figure*}
\begin{center}
\includegraphics[scale=0.3]{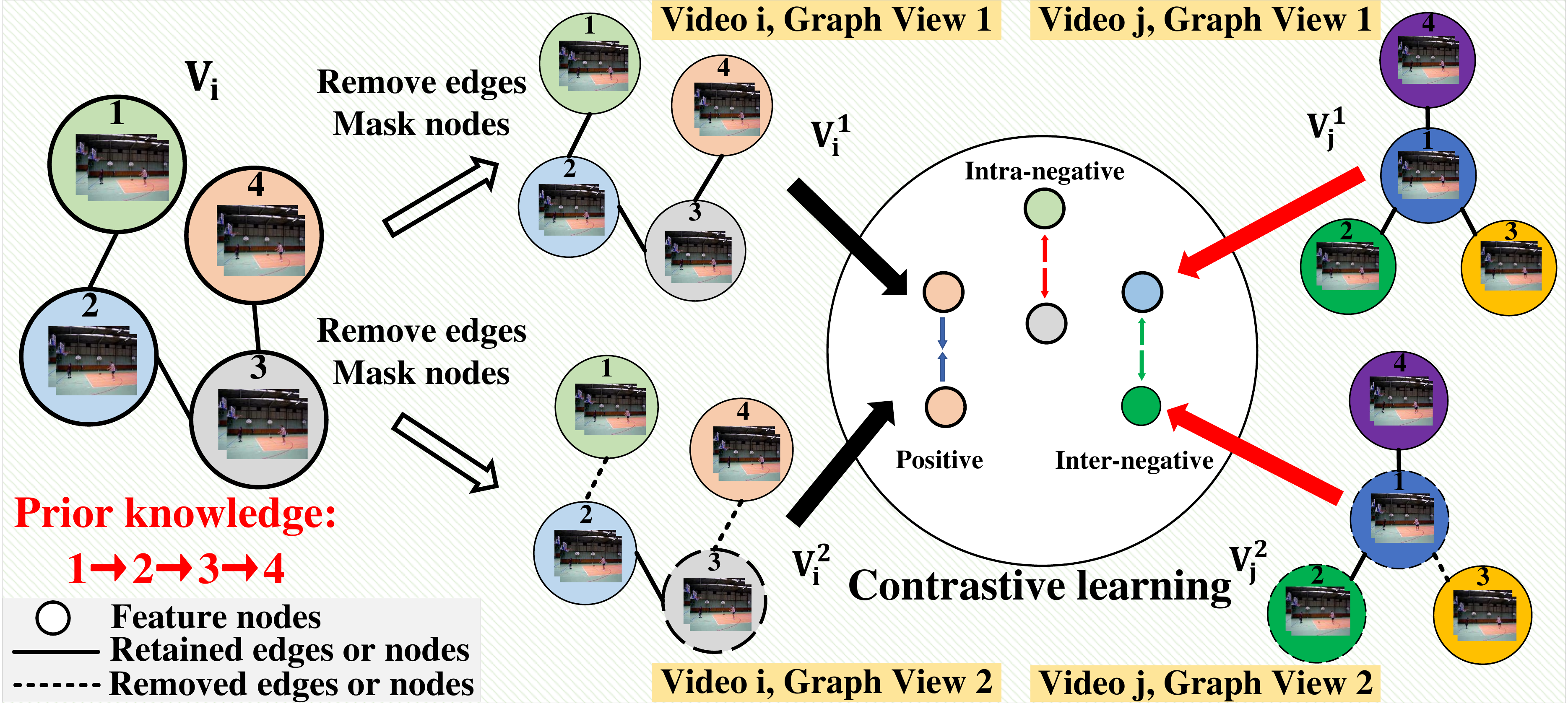}
\end{center}
  \vspace{-10pt}
   \caption{Details of temporal contrastive graph learning module. }
     \vspace{-15pt}
\label{Fig3}
\end{figure*}

For a better presentation, we first introduce several definitions. Given a video $V$, the snippets from this video are composed of continuous frames with the size $c\times l\times h\times w$, where $c$ is the number of channels, $l$ is the number of frames, $h$ and $w$ indicate the height and width of frames. The size of the 3D convolutional kernel is $t\times d\times d$, where $t$ is the temporal length and $d$ is the spatial size. We define an ordered snippet tuples as $\mathbf{S}=\langle s_1, s_2, \cdots, s_n\rangle$, the frame-sets from snippet $s_i $ is denoted as $\mathbf{F}_i=\langle f_1, f_2, \cdots, f_m\rangle$. The subscripts here represent the chronological order. Let $\mathcal{G}=(\mathcal{V},\mathcal{E})$ denote a graph, where $\mathcal{V}=\{v_1,v_2,\cdots,v_N\}$ represents the node set and $\mathcal{E}\in \mathcal{V}\times \mathcal{V}$ represents the edge set. We denote the feature matrix and the adjacency matrix as $\mathbf{X}\in \mathbb{R}^{N\times F}$ and $\mathbf{A}\in \{0,1\}^{N\times N}$, where $\mathbf{x}_i\in \mathbb{R}^F$ is the feature of $v_i$, and $\mathbf{A}_{ij}=1$ if $(v_i,v_j)\in \mathcal{E}$.

\subsection{Sample and Shuffle}
In this stage, we randomly sample consecutive frames (snippets) from the video to construct video snippet tuples. If we sample $N$ snippets from a video, there are $N!$ possible snippet orders. Since the snippet order prediction is purely a proxy task of the TCGL and our focus is the learning of 3D CNNs, we restrict the number of snippets of a video between $3$ to $4$ to alleviate the complexity of the order prediction task, inspired by the previous works \cite{noroozi2016unsupervised,xu2019self,xiao2020explore}. The snippets are sampled uniformly from the video with the interval of $p$ frames. After sampling, the snippets are shuffled to form the snippet tuples $\mathbf{S}=\langle s_1, s_2, \cdots, s_n\rangle$. For each snippet $s_i$, all the frames within are uniformly divided into $m$ frame-sets with equal length, then we get the frame-set $\mathbf{F}_i=\langle f_1, f_2, \cdots, f_m\rangle$ for snippet $s_i$. For snippet tuples, they contain dynamic information and strict chronological relations of a video, which is essentially the long-term temporal dependency of the videos. For the frame-sets within a snippet, the frame-level temporal relation among frames provides us the short-term temporal dependency of the videos. By taking both long-term and short-term temporal dependencies into consideration, we can increase the temporal diversity more comprehensively and precisely.

\subsection{Feature Extraction}
To extract spatio-temporal features, we choose C3D \cite{tran2015learning}, R3D \cite{tran2018closer} and R(2+1)D \cite{tran2018closer} as feature encoders. The same 3D CNNs are used for all snippets and frame-sets, as Figure \ref{Fig2} (b) shows. C3D is an extension from 2D CNNs for spatio-temporal representation learning since it can model the temporal information and dynamics of the videos. C3D network consists of $8$ convolutional layers stacked one by one, with $5$ pooling layers interleaved, and followed by two fully connected layers. The size of all convolutional kernels is $3\times3\times3$, which is validated in previous work \cite{tran2015learning}. R3D is the 3D CNNs with residual connections. R3D block consists of two 3D convolutional layers followed by batch normalization and ReLU layers. The input and output are connected with a residual unit before the ReLU layer. R(2+1)D is similar to R3D. The 3D convolution is decomposed into two separate operations, the one is 2D spatial convolution, and the other is 1D temporal convolution.


\subsection{Temporal Contrastive Graphs}

Due to the effectiveness of graph convolutional network (GCN) \cite{kipf2016semi,zhu2020deep,wu2020comprehensive} in modeling unregular graph-structured relationship, we use it to explore node interaction within each snippet and frame-set for modeling multi-scale temporal dependencies of videos. After obtaining the feature vectors for snippets and frame-sets, we construct two kinds of temporal contrastive graph structures: inter-snippet and intra-snippet temporal contrastive graphs, to increase the temporal diversity of videos, as shown in Figure \ref{Fig2} (c).

To build intra-snippet and inter-snippet temporal contrastive graphs, we take advantage of prior knowledge about the chronological relation and the corresponding feature vectors. To fix notation, we denote intra-snippet and inter-snippet graphs as $\mathbf{G}_{intra}^k=f(\mathbf{X}_{intra}^k,\mathbf{A}_{intra}^k)$ and $\mathbf{G}_{inter}=f(\mathbf{X}_{inter},\mathbf{A}_{inter})$, respectively, where $k=1,\cdots,m$, $m$ is the number of frame-sets in a video snippet. As shown in Figure \ref{Fig3}, the prior knowledge that the correct order of frames in frame-sets, and the correct order of snippets in snippet tuples are already known because our proxy task is video snippet order prediction. Therefore, we can utilize the prior chronological relationship to determine the edges of graphs. For example, in Figure \ref{Fig3}, if we know that snippets (frame-sets) are ranking chronologically $1\rightarrow2\rightarrow3\rightarrow4$, we can connect the temporally related nodes and disconnect the temporally unrelated nodes. Here, we take inter-snippet temporal graph $\mathbf{G}_{inter}$ as an example to clarify our temporal contrastive graph learning method.

For $\mathbf{G}_{inter}$, we randomly remove edges and masking node features to generate two graph views $\mathbf{\widetilde{G}}_{inter}^1$ and $\mathbf{\widetilde{G}}_{inter}^2$, and the node embeddings of two generated views are denoted as $\mathbf{U}=f(\mathbf{\widetilde{X}}_{inter}^1,\mathbf{\widetilde{A}}_{inter}^1)$ and $\mathbf{V}=f(\mathbf{\widetilde{X}}_{inter}^2,\mathbf{\widetilde{A}}_{inter}^2)$. Since different graph views provide different contexts for each node, we corrupt the original graph at both structure and attribute levels to achieve contrastive learning between node embeddings from different views. Therefore, we propose two strategies for generating graph views: removing edges and masking node features.

The edges in the original graph are randomly removed using a random masking matrix $\mathbf{\widetilde{R}}\in\{0,1\}^{N\times N}$, where the entry of $\mathbf{\widetilde{R}}$ is drawn from a Bernoulli distribution $\mathbf{\widetilde{R}}_{ij}\sim \mathcal{B}(1-p_r)$ if $\mathbf{A}_{ij}=1$ for the original graph, and $\mathbf{\widetilde{R}}_{ij}=0$ otherwise. Here $p_r$ denotes the probability of each edge being removed. Then, the resulting adjacency matrix can be computed as follows, where $\circ$ is Hadamard product.
\begin{equation}\label{eq1}
\mathbf{\widetilde{A}}=\mathbf{A}\circ\mathbf{\widetilde{R}}
\end{equation}
In addition, a part of node features is masked with zeros using a random vector $\mathbf{\widetilde{m}}\in\{0,1\}^F$, where each dimension of it is drawn from a Bernoulli distribution $\widetilde{m}_i\sim \mathcal{B}(1-p_m), \forall i$. Then, the generated masked features $\mathbf{\widetilde{X}}$ is calculated as follows:
\begin{equation}\label{eq2}
\mathbf{\widetilde{X}}=[\mathbf{x}_1\circ \mathbf{\widetilde{m}};\mathbf{x}_2\circ \mathbf{\widetilde{m}};\cdots; \mathbf{x}_N\circ \mathbf{\widetilde{m}}]^\top
\end{equation}
where $[\cdot;\cdot]$ is the concatenation operator. We jointly leverage these two strategies to generate graph views.

Inspired by the NCE loss \cite{gutmann2010noise}, we propose a contrastive learning loss that distinguishes embeddings of the same node from these two distinct views from other node embeddings. Given a positive pair, the negative samples come from all other nodes in the two views (inter-view or intra-view). To compute the relationship of embeddings $\mathbf{u}$, $\mathbf{v}$ from two views, we define the relation function $\phi(\mathbf{u},\mathbf{v})=f(g(\mathbf{u}),g(\mathbf{v}))$, where $f$ is the L2 normalized dot product similarity, and $g$ is a non-linear projection with two-layer multi-layer perception. The pairwise contrastive objective for positive pair $(\mathbf{u}_i,\mathbf{v}_i)$ is defined as:
\begin{equation}\label{eq3}
\ell(\mathbf{u}_i,\mathbf{v}_i)=\frac{e^{\frac{\phi(\mathbf{u_i},\mathbf{v_i})}{\tau}}}{e^{\frac{\phi(\mathbf{u_i},\mathbf{v_i})}{\tau}}+\sum\limits_{k=1}^N{\mathbf{\mathbb{I}}_{[k\neq i]}e^{\frac{\phi(\mathbf{u_i},\mathbf{v_k})}{\tau}}}+e^{\frac{\phi(\mathbf{u_i},\mathbf{u_k})}{\tau}}}
\end{equation}
where $\mathbf{\mathbb{I}}_{[k\neq i]}\in\{0,1\}$ is an indicator function that equals to $1$ if $k\neq i$, and $\tau$ is a temperature parameter, which is empirically set to $0.5$. The first term in the denominator represents the positive pairs, the second term represents the inter-view negative pairs, the third term represents the intra-view negative pairs. Since two views are symmetric, the loss for another view is defined similarly for $\ell(\mathbf{v}_i,\mathbf{u}_i)$. The overall contrastive loss for $\mathbf{G}_{inter}$ is defined as follows:
\begin{equation}\label{eq4}
\mathcal{J}_{inter}=\frac{1}{2N}\sum\limits_{i=1}^N[\ell(\mathbf{u}_i,\mathbf{v}_i)+\ell(\mathbf{v}_i,\mathbf{u}_i)]
\end{equation}

The contrastive loss for intra-snippet graphs $\mathbf{G}_{intra}^k ( k=1,\cdots,m)$ can be computed similarly as $\mathbf{G}_{inter}$.
\begin{equation}\label{eq5}
\mathcal{J}_{intra}^k=\frac{1}{2N}\sum\limits_{i=1}^N[\ell^k(\mathbf{u}_i,\mathbf{v}_i)+\ell^k(\mathbf{v}_i,\mathbf{u}_i)]
\end{equation}

Then, the overall temporal contrastive graph loss is defined as follows, where $\alpha$ and $\beta$ are the weights for inter-snippet graph and intra-snippet graph, respectively.
\begin{equation}\label{eq6}
\mathcal{J}_{g}=\alpha\sum\limits_{k=1}^m\mathcal{J}_{intra}^k+\beta\mathcal{J}_{inter}
\end{equation}

\subsection{Adaptive Order Prediction}

\begin{figure}
\begin{center}
\includegraphics[scale=0.5]{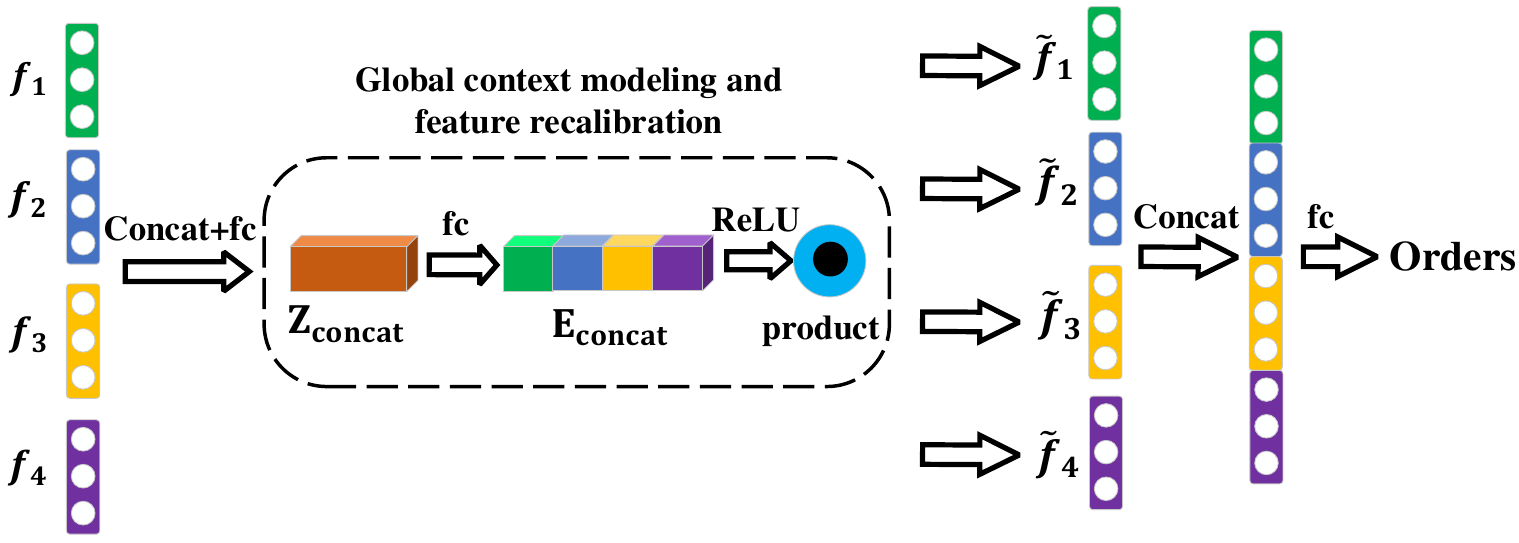}
\end{center}
  \vspace{-10pt}
   \caption{Adaptive snippet order prediction module.}
    \vspace{-15pt}
\label{Fig4}
\end{figure}

We formulate the order prediction task as a classification task using the learned video snippet features from the temporal contrastive graph as the input and the probability distribution of orders as output. Since the features from different video snippets are correlated, we build an adaptive order prediction module that receives features from different video snippets and learns a global context embedding, then this embedding is used to recalibrate the input features from different snippets, as shown in Figure \ref{Fig4}.

To fix notation, we assume that a video is shuffled into $n$ snippets, the snippet features of nodes learned from inter-snippet temporal contrastive graph learning are $\{\mathbf{f}_1,\cdots,\mathbf{f}_n\}$, where $\mathbf{f}_k\in R^{c_k} (k=1,\cdots,n$). To utilize the correlation among these snippets, we concatenate these feature vectors and obtain joint representations through a fully-connected layer:
\begin{equation}\label{eq7}
\mathbf{Z}=\mathbf{W}_{s}[\mathbf{f}_1,\cdots,\mathbf{f}_n]+\mathbf{b}_{s}
\end{equation}
where $[\cdot,\cdot]$ denotes the concatenation operation, $\mathbf{Z}\in \mathbb{R}^{c_{con}}$ denotes the joint representation, $\mathbf{W}_{s}$ and $\mathbf{b}_{s}$ are weights and bias of the fully-connected layer. We choose $c_{con}=\frac{\sum_{k=1}^nc_k}{2n}$ to restrict the model capacity and increase its generalization ability. To make use of the global context information aggregated in the joint representations $\mathbf{Z}_{con}$, we predict excitation signal for it by a fully-connected layer:
\begin{equation}\label{eq8}
\mathbf{E}=\mathbf{W}_{e}\mathbf{Z}+\mathbf{b}_{e}
\end{equation}
where $\mathbf{W}_{e}$ and $\mathbf{b}_{e}$ are weights and biases of the fully-connected layer. After obtaining the excitation signal $\mathbf{E}\in \mathbb{R}^{c}$, we use it to recalibrate the input feature $f_k$ adaptively by a simple gating mechanism:
\begin{equation}\label{eq9}
\mathbf{\widetilde{f}}_k=\delta(\mathbf{E})\odot \mathbf{f}_k
\end{equation}
where $\odot$ is channel-wise product operation for each element in the channel dimension, and $\delta(\cdot)$ is the ReLU function. In this way, we can allow the features of one snippet to recalibrate the features of another snippet while concurrently preserving the correlation among different snippets.

Finally, these refined features $\{\mathbf{\widetilde{f}}_1,\cdots,\mathbf{\widetilde{f}}_n\}$ are concatenated and fed into two-layer perception with soft-max to output the snippet order prediction. The cross-entropy loss is used to measure the correctness of the prediction:
\begin{equation}\label{eq10}
\mathcal{J}_{o}=-\sum_{i=1}^{C}\mathbf{y}_i\textrm{log}(\mathbf{p}_i)
\end{equation}
where $y_i$ and $p_i$ represent the probability that the sample belongs to the order class $i$ in ground-truth and prediction, respectively. $C$ denotes the number of all possible orders.

The overall self-supervised learning loss for TCGL is obtained by combing Eq (\ref{eq6}) and Eq  (\ref{eq10}), where $\lambda_g$ and $\lambda_o$ control the contribution of $\mathcal{J}_{g}$ and $\mathcal{J}_{o}$, respectively.
\begin{equation}\label{eq11}
\mathcal{J}=\lambda_g\mathcal{J}_{g}+\lambda_o\mathcal{J}_{o}
\end{equation}

\section{Experiments}\vspace{-5pt}
In this section, we first elaborate experimental settings, and then conduct ablation studies to analyze the contribution of key components. Finally, the learned 3D CNNs are evaluated on video action recognition and video retrieval tasks, and then compared with state-of-the-art methods.
\subsection{Experimental Setting}
\textbf{Datasets.} We evaluate our method on three action recognition datasets, UCF101 \cite{soomro2012ucf101}, HMDB51 \cite{kuehne2011hmdb} and Kinetics-400 \cite{kay2017kinetics}. UCF101 is collected from websites containing $101$ action classes with $9.5$k videos for training and $3.5$k videos for testing. HMDB51 is collected from various sources with $51$ action classes and $3.4$k videos for training and $1.4$k videos for testing. Kinetics-400 is a large-scale action recognition dataset, which contains 400 action classes and around 306k videos. In this work , we use the training split (around 240k videos) as the pre-training dataset.

\textbf{Network Architecture.} We use PyTorch \cite{paszke2017automatic} to implement the whole framework. For video encoder, C3D, R3D and R(2+1)D are used as backbones, where the kernel size of 3D convolutional layers is set to $3\times3\times3$. The R3D network is implemented with no repetitions in conv\{2-5\}\_x, which results in $9$ convolution layers in total. The C3D network is modified by replacing the two fully connected layers with global spatiotemporal pooling layers. The R(2+1)D network has the same architecture as the R3D network with only 3D kernels decomposed. Dropout layers are applied between fully-connected layers with $p=0.1$. Our GCN for both inter-snippet and intra-snippet graphs consist of one graph convolutional layer with $512$ output channels.

\textbf{Parameters.} Following the settings in \cite{xu2019self,yao2020video}, we set the snippet length of input video as $16$, the interval length is set as $8$, the number of snippets per tuple is $3$, and the number of frame-sets within each snippet is $4$. During training, we randomly split $800$ videos from the training set as the validation set. Video frames are resized to $128\times171$ and then randomly cropped to $112\times112$. We set the parameters $\lambda_g=\lambda_o=1$ to balance the contribution between temporal contrastive graph module and adaptive order prediction module. The weights $\alpha$ and $\beta$ are both set to $1$ according to the ablation study result. To optimize the framework, we use mini-bach stochastic gradient descent with the batchsize $16$, the initial learning rate $0.001$, the momentum $0.9$ and the weight decay $0.0005$. The training process lasts for $300$ epochs and the learning rate is decreased to $0.0001$ after $150$ epochs. To make temporal contrastive graphs sensitive to subtle variance between different graph views, the parameters $p_r$ and $p_m$ for generating graph view 1 are empirically set to $0.2$ and $0.1$, and $p_r=p_m=0$ for generating graph view 2. And the values of $p_r$ and $p_m$ are the same for both inter-snippet and intra-snippet graphs. The model with the lowest validation loss is saved to the best model.

\subsection{Ablation Study}
In this subsection, we conduct ablation studies on the first split of UCF101 with R3D as the backbone, to analyze the contribution of each component of our TCGL.

\textbf{The number of snippets}. The results of R3D on the snippet order prediction task with different number of snippets are shown in Table \ref{Table1}. The prediction accuracy decreases when the number of snippets increases because the difficulty of the prediction task grows when the snippets number increase, which makes the model hard to learn. Therefore, we use $3$ snippets per video to make a compromise between task complexity and prediction accuracy.

\textbf{The number of frame-sets}. Since the snippet length is $16$, the number of frame-sets within each snippet can be $1,2,4,8,16$. When the number is $16$, the frame-set only contains static information without temporal information. When the number is $1$ or $2$, it is hard to model the intra-snippet temporal relationship with too few frame-sets. From Table \ref{Table2}, we can observe that more frame-sets within a snippet will make the intra-snippet temporal modeling more difficult, which degrades the order prediction performance. Therefore, we choose $4$ frame-sets per snippet in the experiments for short-term temporal modeling.

\textbf{The intra-snippet and inter-snippet graphs}. To analyze the contribution of intra-snippet and inter-snippet temporal contrastive graphs, we set different values to $\alpha$ and $\beta$, shown in Table \ref{Table3}. To be noticed, removing intra-snippet graph will degrade the performance significantly even with the inter-snippet graph, which verifies the importance of intra-snippet graphs for modeling short-term temporal dependency. Additionally, when setting the weight values of intra-snippet graph and inter-snippet graph to $1$ and $0.1$, the prediction accuracy is $80.2\%$. While exchanging their weight values, the accuracy drops to $54.9\%$, which validates the importance of intra-snippet graphs in modeling short-term temporal dependency. In addition, the performance of TCGL drops significantly when removing either or both of the graphs. When $\alpha=\beta=1$, the prediction accuracy is the best ($83.0\%$). These results validate that both intra-snippet and inter-snippet temporal contrastive graphs are essential for increasing the temporal diversity of features.

\begin{table}\renewcommand\tabcolsep{5.0pt}\renewcommand\arraystretch{1}
\begin{center}
\begin{tabular}{lccc}
\hline
Model&Snippet Length&Snippets Number&Accuracy\\\hline
R3D&16&2&53.4\\
R3D&16&3&\textbf{80.2}\\
R3D&16&4&61.1\\\hline
\end{tabular}
\end{center}
  \vspace{-10pt}
\caption{Snippet order prediction accuracy (\%) with different number of snippets within each video.}
  \vspace{-10pt}
\label{Table1}
\end{table}

\begin{table}\renewcommand\tabcolsep{4.0pt}\renewcommand\arraystretch{1}
\begin{center}
\begin{tabular}{lccc}
\hline
Model&Snippet Length&Frame-set Number&Accuracy\\\hline
R3D&16&1&54.1\\
R3D&16&2&54.1\\
R3D&16&4&\textbf{80.2}\\
R3D&16&8&63.1\\\hline
\end{tabular}
\end{center}
  \vspace{-10pt}
\caption{Snippet order prediction accuracy (\%) with various number of frame-sets within each snippet.}
  \vspace{-10pt}
\label{Table2}
\end{table}

\begin{table}\renewcommand\tabcolsep{5pt}\renewcommand\arraystretch{1}
\begin{center}
\begin{tabular}{lcccc}
\hline
Model&Intra ($\alpha$)&Inter ($\beta$)&Prediction&Recognition\\\hline
R3D&0&0&55.6&57.3\\
R3D&0&1&55.6&56.0\\
R3D&1&0&76.6&60.9\\
R3D&0.1&1&54.9&56.7\\
R3D&1&0.1&80.2&66.8\\
R3D&1&1&\textbf{83.0}&\textbf{67.6}\\\hline
\end{tabular}
\end{center}
  \vspace{-10pt}
\caption{Snippet order prediction and action recognition accuracy (\%) with different weights ($\alpha$, $\beta$) of intra-/inter- snippet graphs.}
  \vspace{-10pt}
\label{Table3}
\end{table}

\begin{table}\renewcommand\tabcolsep{6.0pt}\renewcommand\arraystretch{1}
\begin{center}
\begin{tabular}{lcccc}
\hline
Method&Model&ASOR&Prediction&Recognition\\\hline
TCGL &R3D&\XSolidBrush&78.4&62.8\\
TCGL &R3D&\Checkmark&\textbf{80.2}&\textbf{67.6}\\\hline
\end{tabular}
\end{center}
  \vspace{-10pt}
\caption{Snippet order prediction and action recognition accuracy (\%) with/without adaptive snippet order prediction (ASOR).}
  \vspace{-15pt}
\label{Table4}
\end{table}

\textbf{The adaptive snippet order prediction}. To analyze the contribution of our proposed adaptive snippet order prediction (ASOR) module, we remove this module and merely feed the concatenated features into multi-layer perception with soft-max to output the final snippet order prediction. It can be observed in Table \ref{Table4} that our TCGL performs better than the TCGL without the ASOR module in both order prediction and action recognition tasks. This verifies that the ASOR module can better utilize relational knowledge among video snippets than simple concatenation.

\textbf{The methods of introducing noises}. To justify our superiority on modeling both feature and topology levels, we make comparisons with three simple graph corruption methods: (1) adding random Gaussian noise; (2) randomly removing edges; (3) randomly removing nodes. The (order prediction, action recognition) accuracies (\%) for them are ($56.1, 51.8$), ($55.2, 53.4$) and ($54.9, 52.7$), respectively, while we achieve ($80.2, 67.6$), justifying the our superiority on modeling both feature and topology levels.

\begin{table}\renewcommand\tabcolsep{1pt}\renewcommand\arraystretch{1}
\begin{center}
\begin{tabular}{lcccc}
\hline
Method&Backbone&Pretrain&UCF101&HMDB51\\\hline
Object Patch \cite{wang2015unsupervised}&AlexNet&UCF101&42.7&15.6 \\
Shuffle \cite{misra2016shuffle}&CaffeNet&UCF101&50.9&19.8 \\
OPN \cite{lee2017unsupervised}&VGG&UCF101&56.3&22.1 \\
Deep RL \cite{buchler2018improving}&CaffeNet&UCF101&58.6&25.0 \\\hline
Random (Baseline)&C3D&UCF101&61.8&24.7\\
Mas \cite{wang2019self}&C3D&UCF101&58.8&32.6 \\
VCOP\cite{xu2019self}&C3D&UCF101&65.6&28.4\\
COP\cite{xiao2020explore}&C3D&UCF101&66.9&31.8\\
PRP \cite{yao2020video}&C3D&UCF101&69.1&34.5\\
ST-puzzle \cite{kim2019self}&C3D&K400&60.6&28.3 \\
Mas \cite{wang2019self}&C3D&K400&61.2&33.4\\
STS \cite{wang2021self}&C3D&K400&71.8&37.8\\
\textbf{TCGL (Ours)}&C3D&UCF101&69.5&35.1\\
\textbf{TCGL (Ours)}&C3D&K400&\textbf{75.2}&\textbf{38.9}\\\hline
Random (Baseline)&R3D&UCF101&54.5&23.4\\
VCOP \cite{xu2019self}&R3D&UCF101&64.9&29.5\\
COP \cite{xiao2020explore}&R3D&UCF101&66.0&28.0\\
TCP \cite{lorre2020temporal}&R3D&UCF101&64.8&34.7\\
PRP \cite{yao2020video}&R3D&UCF101&66.5&29.7\\
ST-puzzle \cite{kim2019self}&R3D&K400&65.8&33.7 \\
DPC \cite{han2019video}&R3D&K400&68.2&34.5\\
TCP \cite{lorre2020temporal}&R3D&K400&70.5&41.1\\
\textbf{TCGL (Ours)}&R3D&UCF101&67.6&30.8\\
\textbf{TCGL (Ours)}&R3D&K400&\textbf{76.8}&\textbf{41.5}\\\hline
Random (Baseline)&R(2+1)D&UCF101&55.8&22.0\\
VCP \cite{luo2020video}&R(2+1)D&UCF101&66.3&32.2\\
VCOP \cite{xu2019self}&R(2+1)D&UCF101&72.4&30.9\\
STS \cite{wang2021self}&R(2+1)D&UCF101&73.6&34.1\\
COP \cite{xiao2020explore}&R(2+1)D&UCF101&74.5&34.8\\
PRP \cite{yao2020video}&R(2+1)D&UCF101&72.1&35.0\\
V-pace \cite{wang2020self}&R(2+1)D&UCF101&75.9&35.9\\
V-pace \cite{wang2020self}&R(2+1)D&K400&77.1&36.6\\
\textbf{TCGL (Ours)}&R(2+1)D&UCF101&74.9&36.2\\
\textbf{TCGL (Ours)}&R(2+1)D&K400&\textbf{77.6}&\textbf{39.7}\\\hline
\end{tabular}
\end{center}
  \vspace{-10pt}
\caption{Comparison with the state-of-the-art self-supervised learning methods on UCF101 and HMDB51 datasets.}
  \vspace{-15pt}
\label{Table5}
\end{table}

\subsection{Action Recognition}
To verify the effectiveness of our TCGL in action recognition, we initialize the backbones with the model pretrained on the first split of UCF101 or the whole K400 training-set, and fine-tune on UCF101 and HMDB51, the fine-tuning stops after $150$ epochs. The features extracted by the backbones are fed into fully-connected layers to obtain the prediction. For testing, we sample $10$ clips for each video to generate clip predictions, and then average these predictions to obtain the final prediction results. The average classification accuracy over three splits is reported and compared with other self-supervised methods in Table \ref{Table5}. The ``Random" means the model is randomly initialized without pre-training. When pre-trained on UCF101, we outperform the current best-performing method PRP \cite{yao2020video}. When pre-trained on K400, we outperform the current best-performing method V-pace \cite{wang2020self}. In addition, we consistently outperform the other state-of-the-art methods and random initialization method for all evaluation metrics. Furthermore, when pre-trained on UCF101, we achieve better accuracies than some K400 pre-trained methods (Mas \cite{wang2019self} and ST-puzzle \cite{kim2019self}). These results validate that our TCGL can effectively increase the temporal diversity of videos and learn discriminative spatio-temporal representations.

\begin{table}\renewcommand\tabcolsep{1pt}\renewcommand\arraystretch{1}
\begin{center}
\begin{tabular}{lccccccc}
\hline
Method&Backbone&Pretrain&T1&T5&T10&T20&T50 \\\hline
Jigsaw \cite{noroozi2016unsupervised}&AlexNet&UCF&19.7&28.5&33.5&40.0&49.4\\
OPN \cite{lee2017unsupervised}&VGG&UCF&19.9&28.7&34.0&40.6&51.6\\
Deep RL \cite{buchler2018improving}&CaffeNet&UCF&25.7&36.2&42.2&49.2&59.5\\
SpeedNet \cite{benaim2020speednet}&S3D-G&K400&13.0&28.1&37.5&49.5&65.0\\\hline
Random &C3D&UCF&16.7&27.5&33.7&41.4&53.0\\
VCOP \cite{xu2019self}&C3D&UCF&12.5&29.0&39.0&50.6&66.9\\
PRP \cite{yao2020video}&C3D&UCF&23.2&38.1&46.0&55.7&68.4\\
V-pace \cite{wang2020self}&C3D&UCF&20.0&37.4&46.9&58.5&73.1\\
\textbf{TCGL (Ours)}&C3D&UCF&22.5&40.7&49.8&59.9&73.3\\
\textbf{TCGL (Ours)}&C3D&K400&\textbf{23.6}&\textbf{41.2}&\textbf{50.1}&\textbf{60.4}&\textbf{74.2}\\\hline
Random &R3D&UCF&9.9&18.9&26.0&35.5&51.9\\
VCOP \cite{xu2019self}&R3D&UCF&14.1&30.3&40.4&51.1&66.5\\
PRP \cite{yao2020video}&R3D&UCF&22.8&38.5&46.7&55.2&69.1\\
V-pace \cite{wang2020self}&R3D&UCF&19.9&36.2&46.1&55.6&69.2\\
\textbf{TCGL (Ours)}&R3D&UCF&23.3&39.6&48.4&58.8&72.4\\
\textbf{TCGL (Ours)}&R3D&K400&\textbf{23.9}&\textbf{43.0}&\textbf{53.0}&\textbf{62.9}&\textbf{75.7}\\\hline
Random &R(2+1)D &UCF&10.6&20.7&27.4&37.4&53.1\\
VCOP \cite{xu2019self}&R(2+1)D &UCF&10.7&25.9&35.4&47.3&63.9\\
PRP \cite{yao2020video}&R(2+1)D &UCF&20.3&34.0&41.9&51.7&64.2\\
V-pace \cite{wang2020self}&R(2+1)D &UCF&17.9&34.3&44.6&55.5&72.0\\
\textbf{TCGL (Ours)}&R(2+1)D &UCF&20.6&36.2&45.5&56.2&72.1\\
\textbf{TCGL (Ours)}&R(2+1)D &K400&\textbf{21.9}&\textbf{40.2}&\textbf{49.6}&\textbf{59.7}&\textbf{73.1}\\\hline
\end{tabular}
\end{center}
  \vspace{-10pt}
\caption{Video retrieval result (\%) on UCF101. `T' denotes `Top'.}
\vspace{-5pt}
\label{Table6}
\end{table}\vspace{-10pt}

\begin{table}\renewcommand\tabcolsep{1pt}\renewcommand\arraystretch{1}
\begin{center}
\begin{tabular}{lccccccc}
\hline
Method&Backbone&Pretrain&T1&T5&T10&T20&T50 \\\hline
Random &C3D&UCF&7.4&20.5&31.9&44.5&66.3\\
VCOP \cite{xu2019self}&C3D&UCF&7.4&22.6&34.4&48.5&70.1\\
PRP \cite{yao2020video}&C3D&UCF&10.5&27.2&40.4&56.2&75.9\\
V-pace \cite{wang2020self}&C3D&UCF&8.0&25.2&37.8&54.4&77.5\\
\textbf{TCGL (Ours)}&C3D&UCF&10.7&28.6&41.1&57.9&77.7\\
\textbf{TCGL (Ours)}&C3D&K400&\textbf{12.3}&\textbf{30.4}&\textbf{42.9}&\textbf{59.1}&\textbf{79.2}\\\hline
Random &R3D&UCF&6.7&18.3&28.3&43.1&67.9\\
VCOP \cite{xu2019self}&R3D&UCF&7.6&22.9&34.4&48.8&68.9\\
PRP \cite{yao2020video}&R3D&UCF&8.2&25.8&38.5&53.3&75.9\\
V-pace \cite{wang2020self}&R3D&UCF&8.2&24.2&37.3&53.3&74.5\\
\textbf{TCGL (Ours)}&R3D&UCF&10.9&29.5&42.9&56.9&76.8\\
\textbf{TCGL (Ours)}&R3D&K400&\textbf{11.2}&\textbf{30.6}&\textbf{43.8}&\textbf{58.1}&\textbf{78.0}\\\hline
Random &R(2+1)D &UCF&4.5&14.8&23.4&38.9&63.0\\
VCOP \cite{xu2019self}&R(2+1)D &UCF&5.7&19.5&30.7&45.8&67.0\\
PRP \cite{yao2020video}&R(2+1)D &UCF&8.2&25.3&36.2&51.0&73.0\\
V-pace \cite{wang2020self}&R(2+1)D &UCF&10.1&24.6&37.6&54.4&77.1\\
\textbf{TCGL (Ours)}&R(2+1)D &UCF&11.1&30.4&43.0&56.5&77.4\\
\textbf{TCGL (Ours)}&R(2+1)D &K400&\textbf{13.2}&\textbf{33.5}&\textbf{46.4}&\textbf{59.3}&\textbf{80.2}\\\hline
\end{tabular}
\end{center}
\vspace{-10pt}
\caption{Video retrieval result  (\%) on HMDB51.`T' denotes `Top'. }
 \vspace{-15pt}
\label{Table7}
\end{table}\vspace{10pt}

\subsection{Video Retrieval}
\begin{figure}
\begin{center}
\includegraphics[scale=0.73]{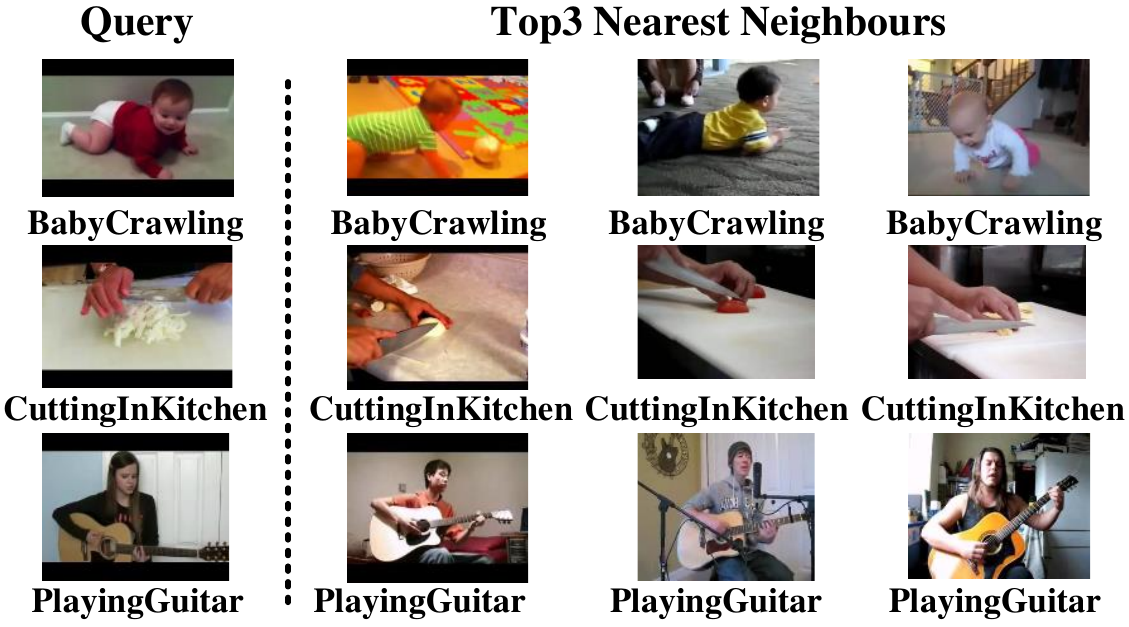}
\end{center}
  \vspace{-10pt}
   \caption{Video retrieval results with TCGL representations.}
     \vspace{-10pt}
\label{Fig5}
\end{figure}

\begin{figure}
\begin{center}
\includegraphics[scale=0.6]{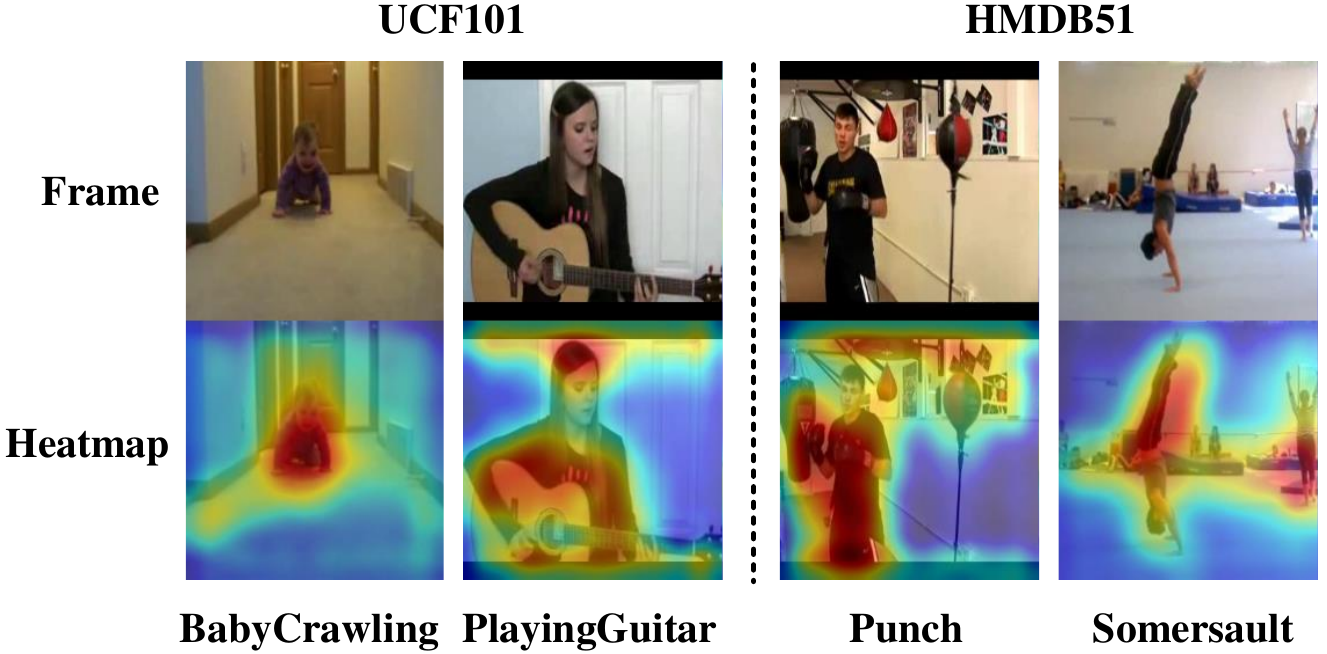}
\end{center}
  \vspace{-10pt}
   \caption{Activation maps on UCF101 and HMDB51 datasets.}
    \vspace{-15pt}
\label{Fig6}
\end{figure}

To further verify the effectiveness of our TCGL in video retrieval, we test our TCGL on the nearest-neighbor video retrieval. Since the video retrieval task is conducted with features extracted by the backbone network without fine-tuning, its performance largely relies upon the representative capacity of self-supervised model. The experiment is conducted on the first split of UCF101 or the whole training-set of K400, following the protocol in \cite{xu2019self,yao2020video}. In video retrieval, we extract video features from the backbone pre-trained by TCGL. Each video in the testing set is used to query $k$ nearest videos from the training set using the cosine distance. When the class of a test video appears in the classes of $k$ nearest training videos, it is considered as the correct predicted video. We show top-1, top-5, top-10, top-20, and top-50 retrieval accuracies on UCF101 and HMDB51 datasets, and compare our method with other self-supervised methods, as shown in Table \ref{Table6} and \ref{Table7}. For all backbones, our TCGL outperforms the state-of-the-art methods on nearly all evaluation metrics by substantial margins. Figure \ref{Fig5} visualizes a query video snippet and its top-3 nearest neighbors from the UCF101 training set using the TCGL embedding. It can be observed that the representation learned by the TCGL has the ability to retrieve videos with the same semantic meaning. To have a better understanding of what TCGL learns, we follow the Class Activation Map \cite{zhou2016learning} to visualize the spatio-temporal regions, as shown in Figure \ref{Fig6}. These examples exhibit a strong correlation between highly activated regions and the dominant movement in the scene. This validates that our TCGL can learn discriminative temporal representations for videos.

\vspace{-5pt}


\section{Conclusion}

In this paper, we propose a novel Temporal Contrastive Graph Learning (TCGL) approach for self-supervised video representation learning. With inter-intra snippet graph contrastive learning strategy and adaptive video snippet order prediction task, the temporal diversity and multi-scale temporal dependency can be well discovered. The proposed TCGL is applied to video action recognition and video retrieval tasks with three kinds of 3D CNNs. Extensive experiments demonstrate the superiority of our TCGL over the state-of-the-art methods on large-scale benchmarks.

\noindent\textbf{Acknowledgements. }This work was supported in part by the National Natural Science Foundation of China under Grant 62002395, the China Postdoctoral Science Foundation funded project under Grant 2020M672966, and the Fundamental Research Funds for the Central Universities under Grant 20lgpy131.

{\small
\bibliographystyle{ieee_fullname}
\bibliography{egbib}

\begin{thebibliography}{10}\itemsep=-1pt

\bibitem{benaim2020speednet}
Sagie Benaim, Ariel Ephrat, Oran Lang, Inbar Mosseri, William~T Freeman,
  Michael Rubinstein, Michal Irani, and Tali Dekel.
\newblock Speednet: Learning the speediness in videos.
\newblock In {\em Proceedings of the IEEE/CVF Conference on Computer Vision and
  Pattern Recognition}, pages 9922--9931, 2020.

\bibitem{buchler2018improving}
Uta Buchler, Biagio Brattoli, and Bjorn Ommer.
\newblock Improving spatiotemporal self-supervision by deep reinforcement
  learning.
\newblock In {\em Proceedings of the European conference on computer vision
  (ECCV)}, pages 770--786, 2018.

\bibitem{DBLP:conf/icml/ChenK0H20}
Ting Chen, Simon Kornblith, Mohammad Norouzi, and Geoffrey~E. Hinton.
\newblock A simple framework for contrastive learning of visual
  representations.
\newblock In {\em Proceedings of the 37th International Conference on Machine
  Learning,}, pages 1597--1607, 2020.

\bibitem{chen2019graph}
Yunpeng Chen, Marcus Rohrbach, Zhicheng Yan, Yan Shuicheng, Jiashi Feng, and
  Yannis Kalantidis.
\newblock Graph-based global reasoning networks.
\newblock In {\em Proceedings of the IEEE/CVF Conference on Computer Vision and
  Pattern Recognition}, pages 433--442, 2019.

\bibitem{cornsweet2012visual}
Tom Cornsweet.
\newblock {\em Visual perception}.
\newblock Academic press, 2012.

\bibitem{deng2009imagenet}
Jia Deng, Wei Dong, Richard Socher, Li-Jia Li, Kai Li, and Li Fei-Fei.
\newblock Imagenet: A large-scale hierarchical image database.
\newblock In {\em 2009 IEEE conference on computer vision and pattern
  recognition}, pages 248--255. Ieee, 2009.

\bibitem{doersch2015unsupervised}
Carl Doersch, Abhinav Gupta, and Alexei~A Efros.
\newblock Unsupervised visual representation learning by context prediction.
\newblock In {\em Proceedings of the IEEE international conference on computer
  vision}, pages 1422--1430, 2015.

\bibitem{fernando2017self}
Basura Fernando, Hakan Bilen, Efstratios Gavves, and Stephen Gould.
\newblock Self-supervised video representation learning with odd-one-out
  networks.
\newblock In {\em Proceedings of the IEEE conference on computer vision and
  pattern recognition}, pages 3636--3645, 2017.

\bibitem{goyal2017something}
Raghav Goyal, Samira~Ebrahimi Kahou, Vincent Michalski, Joanna Materzynska,
  Susanne Westphal, Heuna Kim, Valentin Haenel, Ingo Fruend, Peter Yianilos,
  Moritz Mueller-Freitag, et~al.
\newblock The" something something" video database for learning and evaluating
  visual common sense.
\newblock In {\em ICCV}, volume~1, page~5, 2017.

\bibitem{grill2020bootstrap}
Jean-Bastien Grill, Florian Strub, Florent Altch{\'e}, Corentin Tallec,
  Pierre~H Richemond, Elena Buchatskaya, Carl Doersch, Bernardo~Avila Pires,
  Zhaohan~Daniel Guo, Mohammad~Gheshlaghi Azar, et~al.
\newblock Bootstrap your own latent: A new approach to self-supervised
  learning.
\newblock {\em arXiv preprint arXiv:2006.07733}, 2020.

\bibitem{gutmann2010noise}
Michael Gutmann and Aapo Hyv{\"a}rinen.
\newblock Noise-contrastive estimation: A new estimation principle for
  unnormalized statistical models.
\newblock In {\em Proceedings of the Thirteenth International Conference on
  Artificial Intelligence and Statistics}, pages 297--304, 2010.

\bibitem{hafidi2020graphcl}
Hakim Hafidi, Mounir Ghogho, Philippe Ciblat, and Ananthram Swami.
\newblock Graphcl: Contrastive self-supervised learning of graph
  representations.
\newblock {\em arXiv preprint arXiv:2007.08025}, 2020.

\bibitem{han2019video}
Tengda Han, Weidi Xie, and Andrew Zisserman.
\newblock Video representation learning by dense predictive coding.
\newblock In {\em Proceedings of the IEEE International Conference on Computer
  Vision Workshops}, pages 0--0, 2019.

\bibitem{hans2020visual}
J Hans and Johannes~RM Cruysberg.
\newblock The visual system.
\newblock In {\em Clinical Neuroanatomy}, pages 409--453. Springer, 2020.

\bibitem{he2020momentum}
Kaiming He, Haoqi Fan, Yuxin Wu, Saining Xie, and Ross Girshick.
\newblock Momentum contrast for unsupervised visual representation learning.
\newblock In {\em Proceedings of the IEEE/CVF Conference on Computer Vision and
  Pattern Recognition}, pages 9729--9738, 2020.

\bibitem{ji2020action}
Jingwei Ji, Ranjay Krishna, Li Fei-Fei, and Juan~Carlos Niebles.
\newblock Action genome: Actions as compositions of spatio-temporal scene
  graphs.
\newblock In {\em Proceedings of the IEEE/CVF Conference on Computer Vision and
  Pattern Recognition}, pages 10236--10247, 2020.

\bibitem{kay2017kinetics}
Will Kay, Joao Carreira, Karen Simonyan, Brian Zhang, Chloe Hillier, Sudheendra
  Vijayanarasimhan, Fabio Viola, Tim Green, Trevor Back, Paul Natsev, et~al.
\newblock The kinetics human action video dataset.
\newblock {\em arXiv preprint arXiv:1705.06950}, 2017.

\bibitem{kim2019self}
Dahun Kim, Donghyeon Cho, and In~So Kweon.
\newblock Self-supervised video representation learning with space-time cubic
  puzzles.
\newblock In {\em Proceedings of the AAAI Conference on Artificial
  Intelligence}, volume~33, pages 8545--8552, 2019.

\bibitem{kipf2016semi}
Thomas~N Kipf and Max Welling.
\newblock Semi-supervised classification with graph convolutional networks.
\newblock {\em arXiv preprint arXiv:1609.02907}, 2016.

\bibitem{klaser2008spatio}
Alexander Klaser, Marcin Marsza{\l}ek, and Cordelia Schmid.
\newblock A spatio-temporal descriptor based on 3d-gradients.
\newblock In {\em BMVC 2008-19th British Machine Vision Conference}, pages
  275--1, 2008.

\bibitem{krizhevsky2012imagenet}
Alex Krizhevsky, Ilya Sutskever, and Geoffrey~E Hinton.
\newblock Imagenet classification with deep convolutional neural networks.
\newblock In {\em Advances in neural information processing systems}, pages
  1097--1105, 2012.

\bibitem{kuehne2011hmdb}
Hildegard Kuehne, Hueihan Jhuang, Est{\'\i}baliz Garrote, Tomaso Poggio, and
  Thomas Serre.
\newblock Hmdb: a large video database for human motion recognition.
\newblock In {\em 2011 International Conference on Computer Vision}, pages
  2556--2563. IEEE, 2011.

\bibitem{laptev2005space}
Ivan Laptev.
\newblock On space-time interest points.
\newblock {\em International journal of computer vision}, 64(2-3):107--123,
  2005.

\bibitem{larsson2017colorization}
Gustav Larsson, Michael Maire, and Gregory Shakhnarovich.
\newblock Colorization as a proxy task for visual understanding.
\newblock In {\em Proceedings of the IEEE Conference on Computer Vision and
  Pattern Recognition}, pages 6874--6883, 2017.

\bibitem{lee2017unsupervised}
Hsin-Ying Lee, Jia-Bin Huang, Maneesh Singh, and Ming-Hsuan Yang.
\newblock Unsupervised representation learning by sorting sequences.
\newblock In {\em Proceedings of the IEEE International Conference on Computer
  Vision}, pages 667--676, 2017.

\bibitem{lin2019tsm}
Ji Lin, Chuang Gan, and Song Han.
\newblock Tsm: Temporal shift module for efficient video understanding.
\newblock In {\em Proceedings of the IEEE International Conference on Computer
  Vision}, pages 7083--7093, 2019.

\bibitem{liu2018hierarchically}
Yang Liu, Zhaoyang Lu, Jing Li, and Tao Yang.
\newblock Hierarchically learned view-invariant representations for cross-view
  action recognition.
\newblock {\em IEEE Transactions on Circuits and Systems for Video Technology},
  29(8):2416--2430, 2018.

\bibitem{liu2018global}
Yang Liu, Zhaoyang Lu, Jing Li, Tao Yang, and Chao Yao.
\newblock Global temporal representation based cnns for infrared action
  recognition.
\newblock {\em IEEE Signal Processing Letters}, 25(6):848--852, 2018.

\bibitem{liu2019deep}
Yang Liu, Zhaoyang Lu, Jing Li, Tao Yang, and Chao Yao.
\newblock Deep image-to-video adaptation and fusion networks for action
  recognition.
\newblock {\em IEEE Transactions on Image Processing}, 29:3168--3182, 2019.

\bibitem{livingstone1988segregation}
Margaret Livingstone and David Hubel.
\newblock Segregation of form, color, movement, and depth: anatomy, physiology,
  and perception.
\newblock {\em Science}, 240(4853):740--749, 1988.

\bibitem{lorre2020temporal}
Guillaume Lorre, Jaonary Rabarisoa, Astrid Orcesi, Samia Ainouz, and Stephane
  Canu.
\newblock Temporal contrastive pretraining for video action recognition.
\newblock In {\em Proceedings of the IEEE/CVF Winter Conference on Applications
  of Computer Vision}, pages 662--670, 2020.

\bibitem{luo2020video}
Dezhao Luo, Chang Liu, Yu Zhou, Dongbao Yang, Can Ma, Qixiang Ye, and Weiping
  Wang.
\newblock Video cloze procedure for self-supervised spatio-temporal learning.
\newblock In {\em Proceedings of the AAAI Conference on Artificial
  Intelligence}, volume~34, pages 11701--11708, 2020.

\bibitem{milner2006visual}
David Milner and Mel Goodale.
\newblock {\em The visual brain in action}, volume~27.
\newblock OUP Oxford, 2006.

\bibitem{misra2016shuffle}
Ishan Misra, C~Lawrence Zitnick, and Martial Hebert.
\newblock Shuffle and learn: unsupervised learning using temporal order
  verification.
\newblock In {\em European Conference on Computer Vision}, pages 527--544,
  2016.

\bibitem{nguyen2014stap}
Tam~V Nguyen, Zheng Song, and Shuicheng Yan.
\newblock Stap: Spatial-temporal attention-aware pooling for action
  recognition.
\newblock {\em IEEE Transactions on Circuits and Systems for Video Technology},
  25(1):77--86, 2014.

\bibitem{noroozi2016unsupervised}
Mehdi Noroozi and Paolo Favaro.
\newblock Unsupervised learning of visual representations by solving jigsaw
  puzzles.
\newblock In {\em European Conference on Computer Vision}, pages 69--84, 2016.

\bibitem{paszke2017automatic}
Adam Paszke, Sam Gross, Soumith Chintala, Gregory Chanan, Edward Yang, Zachary
  DeVito, Zeming Lin, Alban Desmaison, Luca Antiga, and Adam Lerer.
\newblock Automatic differentiation in pytorch.
\newblock 2017.

\bibitem{pathak2016context}
Deepak Pathak, Philipp Krahenbuhl, Jeff Donahue, Trevor Darrell, and Alexei~A
  Efros.
\newblock Context encoders: Feature learning by inpainting.
\newblock In {\em Proceedings of the IEEE conference on computer vision and
  pattern recognition}, pages 2536--2544, 2016.

\bibitem{peng2016bag}
Xiaojiang Peng, Limin Wang, Xingxing Wang, and Yu Qiao.
\newblock Bag of visual words and fusion methods for action recognition:
  Comprehensive study and good practice.
\newblock {\em Computer Vision and Image Understanding}, 150:109--125, 2016.

\bibitem{qiu2020gcc}
Jiezhong Qiu, Qibin Chen, Yuxiao Dong, Jing Zhang, Hongxia Yang, Ming Ding,
  Kuansan Wang, and Jie Tang.
\newblock Gcc: Graph contrastive coding for graph neural network pre-training.
\newblock In {\em Proceedings of the 26th ACM SIGKDD International Conference
  on Knowledge Discovery \& Data Mining}, pages 1150--1160, 2020.

\bibitem{simonyan2014two}
Karen Simonyan and Andrew Zisserman.
\newblock Two-stream convolutional networks for action recognition in videos.
\newblock In {\em Advances in neural information processing systems}, pages
  568--576, 2014.

\bibitem{soomro2012ucf101}
Khurram Soomro, Amir~Roshan Zamir, and Mubarak Shah.
\newblock Ucf101: A dataset of 101 human actions classes from videos in the
  wild.
\newblock {\em arXiv preprint arXiv:1212.0402}, 2012.

\bibitem{tao2020self}
Li Tao, Xueting Wang, and Toshihiko Yamasaki.
\newblock Self-supervised video representation learning using inter-intra
  contrastive framework.
\newblock In {\em Proceedings of the 28th ACM International Conference on
  Multimedia}, pages 2193--2201, 2020.

\bibitem{tran2015learning}
Du Tran, Lubomir Bourdev, Rob Fergus, Lorenzo Torresani, and Manohar Paluri.
\newblock Learning spatiotemporal features with 3d convolutional networks.
\newblock In {\em Proceedings of the IEEE international conference on computer
  vision}, pages 4489--4497, 2015.

\bibitem{tran2018closer}
Du Tran, Heng Wang, Lorenzo Torresani, Jamie Ray, Yann LeCun, and Manohar
  Paluri.
\newblock A closer look at spatiotemporal convolutions for action recognition.
\newblock In {\em Proceedings of the IEEE conference on Computer Vision and
  Pattern Recognition}, pages 6450--6459, 2018.

\bibitem{van1994neural}
David~C Van~Essen and Jack~L Gallant.
\newblock Neural mechanisms of form and motion processing in the primate visual
  system.
\newblock {\em Neuron}, 13(1):1--10, 1994.

\bibitem{velivckovic2018graph}
Petar Veli{\v{c}}kovi{\'c}, Guillem Cucurull, Arantxa Casanova, Adriana Romero,
  Pietro Li{\`o}, and Yoshua Bengio.
\newblock Graph attention networks.
\newblock In {\em International Conference on Learning Representations}, 2018.

\bibitem{wang2019adaptively}
Guangrun Wang, Keze Wang, and Liang Lin.
\newblock Adaptively connected neural networks.
\newblock In {\em Proceedings of the IEEE/CVF Conference on Computer Vision and
  Pattern Recognition}, pages 1781--1790, 2019.

\bibitem{wang2013dense}
Heng Wang, Alexander Kl{\"a}ser, Cordelia Schmid, and Cheng-Lin Liu.
\newblock Dense trajectories and motion boundary descriptors for action
  recognition.
\newblock {\em International journal of computer vision}, 103(1):60--79, 2013.

\bibitem{wang2013action}
Heng Wang and Cordelia Schmid.
\newblock Action recognition with improved trajectories.
\newblock In {\em Proceedings of the IEEE international conference on computer
  vision}, pages 3551--3558, 2013.

\bibitem{wang2021self}
Jiangliu Wang, Jianbo Jiao, Linchao Bao, Shengfeng He, Wei Liu, and Yun-Hui
  Liu.
\newblock Self-supervised video representation learning by uncovering
  spatio-temporal statistics.
\newblock {\em IEEE Transactions on Pattern Analysis and Machine Intelligence},
  2021.

\bibitem{wang2019self}
Jiangliu Wang, Jianbo Jiao, Linchao Bao, Shengfeng He, Yunhui Liu, and Wei Liu.
\newblock Self-supervised spatio-temporal representation learning for videos by
  predicting motion and appearance statistics.
\newblock In {\em Proceedings of the IEEE Conference on Computer Vision and
  Pattern Recognition}, pages 4006--4015, 2019.

\bibitem{wang2020self}
Jiangliu Wang, Jianbo Jiao, and Yun-Hui Liu.
\newblock Self-supervised video representation learning by pace prediction.
\newblock In {\em European Conference on Computer Vision}, pages 504--521.
  Springer, 2020.

\bibitem{wang2015action}
Limin Wang, Yu Qiao, and Xiaoou Tang.
\newblock Action recognition with trajectory-pooled deep-convolutional
  descriptors.
\newblock In {\em Proceedings of the IEEE conference on computer vision and
  pattern recognition}, pages 4305--4314, 2015.

\bibitem{wang2018temporal}
Limin Wang, Yuanjun Xiong, Zhe Wang, Yu Qiao, Dahua Lin, Xiaoou Tang, and Luc
  Van~Gool.
\newblock Temporal segment networks for action recognition in videos.
\newblock {\em IEEE transactions on pattern analysis and machine intelligence},
  41(11):2740--2755, 2018.

\bibitem{wang2018non}
Xiaolong Wang, Ross Girshick, Abhinav Gupta, and Kaiming He.
\newblock Non-local neural networks.
\newblock In {\em Proceedings of the IEEE conference on computer vision and
  pattern recognition}, pages 7794--7803, 2018.

\bibitem{wang2015unsupervised}
Xiaolong Wang and Abhinav Gupta.
\newblock Unsupervised learning of visual representations using videos.
\newblock In {\em Proceedings of the IEEE international conference on computer
  vision}, pages 2794--2802, 2015.

\bibitem{wang2018videos}
Xiaolong Wang and Abhinav Gupta.
\newblock Videos as space-time region graphs.
\newblock In {\em Proceedings of the European conference on computer vision
  (ECCV)}, pages 399--417, 2018.

\bibitem{wu2020comprehensive}
Zonghan Wu, Shirui Pan, Fengwen Chen, Guodong Long, Chengqi Zhang, and S~Yu
  Philip.
\newblock A comprehensive survey on graph neural networks.
\newblock {\em IEEE Transactions on Neural Networks and Learning Systems},
  2020.

\bibitem{xiao2020explore}
Jun Xiao, Lin Li, Dejing Xu, Chengjiang Long, Jian Shao, Shifeng Zhang,
  Shiliang Pu, and Yueting Zhuang.
\newblock Explore video clip order with self-supervised and curriculum learning
  for video applications.
\newblock {\em IEEE Transactions on Multimedia}, 2020.

\bibitem{xu2019self}
Dejing Xu, Jun Xiao, Zhou Zhao, Jian Shao, Di Xie, and Yueting Zhuang.
\newblock Self-supervised spatiotemporal learning via video clip order
  prediction.
\newblock In {\em Proceedings of the IEEE Conference on Computer Vision and
  Pattern Recognition}, pages 10334--10343, 2019.

\bibitem{yao2020video}
Yuan Yao, Chang Liu, Dezhao Luo, Yu Zhou, and Qixiang Ye.
\newblock Video playback rate perception for self-supervised spatio-temporal
  representation learning.
\newblock In {\em Proceedings of the IEEE/CVF Conference on Computer Vision and
  Pattern Recognition}, pages 6548--6557, 2020.

\bibitem{zhang2020temporal}
Jingran Zhang, Fumin Shen, Xing Xu, and Heng~Tao Shen.
\newblock Temporal reasoning graph for activity recognition.
\newblock {\em IEEE Transactions on Image Processing}, 29:5491--5506, 2020.

\bibitem{zhang2020deep}
Ziwei Zhang, Peng Cui, and Wenwu Zhu.
\newblock Deep learning on graphs: A survey.
\newblock {\em IEEE Transactions on Knowledge and Data Engineering}, 2020.

\bibitem{zhao2018psanet}
Hengshuang Zhao, Yi Zhang, Shu Liu, Jianping Shi, Chen~Change Loy, Dahua Lin,
  and Jiaya Jia.
\newblock Psanet: Point-wise spatial attention network for scene parsing.
\newblock In {\em Proceedings of the European Conference on Computer Vision
  (ECCV)}, pages 267--283, 2018.

\bibitem{zhou2018temporal}
Bolei Zhou, Alex Andonian, Aude Oliva, and Antonio Torralba.
\newblock Temporal relational reasoning in videos.
\newblock In {\em Proceedings of the European Conference on Computer Vision
  (ECCV)}, pages 803--818, 2018.

\bibitem{zhou2016learning}
Bolei Zhou, Aditya Khosla, Agata Lapedriza, Aude Oliva, and Antonio Torralba.
\newblock Learning deep features for discriminative localization.
\newblock In {\em Proceedings of the IEEE conference on computer vision and
  pattern recognition}, pages 2921--2929, 2016.

\bibitem{zhu2020deep}
Yanqiao Zhu, Yichen Xu, Feng Yu, Qiang Liu, Shu Wu, and Liang Wang.
\newblock Deep graph contrastive representation learning.
\newblock {\em arXiv preprint arXiv:2006.04131}, 2020.

\bibitem{zhuo2019explainable}
Tao Zhuo, Zhiyong Cheng, Peng Zhang, Yongkang Wong, and Mohan Kankanhalli.
\newblock Explainable video action reasoning via prior knowledge and state
  transitions.
\newblock In {\em Proceedings of the 27th ACM International Conference on
  Multimedia}, pages 521--529, 2019.

\end{thebibliography}
}

\end{document}